\documentclass[10pt,twocolumn,letterpaper]{article}

\usepackage{cvpr}              %

\usepackage[table, dvipsnames]{xcolor}
\usepackage{color}
\usepackage{booktabs}
\usepackage{makecell}
\usepackage{multirow}
\usepackage[export]{adjustbox}
\usepackage{mathtools}
\usepackage{comment}
\usepackage{tabularray}
\usepackage{enumitem}

\newcommand{\todo}[1]{{\color{red}#1}}

\definecolor{turquoise}{cmyk}{0.65,0,0.1,0.3}
\definecolor{purple}{rgb}{0.65,0,0.65}
\definecolor{dark_green}{rgb}{0, 0.5, 0}
\definecolor{orange}{rgb}{0.8, 0.6, 0.2}
\definecolor{red}{rgb}{0.8, 0.2, 0.2}
\definecolor{darkred}{rgb}{0.6, 0.1, 0.05}
\definecolor{blueish}{rgb}{0.0, 0.3, .6}
\definecolor{light_gray}{rgb}{0.7, 0.7, .7}
\definecolor{pink}{rgb}{0.9, 0, 0.6}
\definecolor{greyblue}{rgb}{0.25, 0.25, 1}
\definecolor{teal}{rgb}{0.0, 0.4, 0.4}
\definecolor{chocolate}{rgb}{1.0, 0.4, 0.0}

\newcommand{\at}[1]{{\color[rgb]{.5,.5,1}#1}}

\newcommand{\ws}[1]{{\color{dark_green}#1}}

\newcommand{\cl}[1]{{\color{orange}#1}}

\DeclareMathOperator*{\argmin}{arg\,min}

\newcommand{\loss}[1]{\mathcal{L}_\text{#1}}

\newcommand{\param}{\boldsymbol{\theta}}
\newcommand{\expect}{\mathbb{E}}
\newcommand{\real}{\mathbb{R}}

\newcommand{\point}{\mathbf{p}}
\newcommand{\points}{\mathbf{P}}

\newcommand{\query}{\mathbf{q}}

\newcommand{\x}{\mathbf{x}}

\newcommand{\features}{\mathbf{F}}
\newcommand{\feature}{\mathbf{f}}

\newcommand{\neighbor}{\mathcal{N}}

\newcommand{\backbone}{\mathcal{F}}
\newcommand{\aggregation}{\mathcal{A}}
\newcommand{\mlp}{\mathcal{M}}
\newcommand{\pointnet}{\mathcal{E}}
\newcommand{\distancefield}{\mathcal{D}}
\newcommand{\distance}{\mathbf{d}}

\newcommand{\synthetic}{\texttt{SyntheticRoom}~\cite{peng2020convoccnet}\xspace}
\newcommand{\scannet}{\texttt{ScanNet}~\cite{dai2017scannet}\xspace}
\newcommand{\carla}{\texttt{CARLA}~\cite{dosovitskiy2017carla}\xspace}
\newcommand{\scenenn}{\texttt{SceneNN}~\cite{hua2016scenenn}\xspace}
\newcommand{\nksr}{NKSR~\cite{huang2023neural}\xspace}

\definecolor{ours}{rgb}{0.7, 0.7, 0.7} %
\definecolor{1st}{rgb}{0.9, 0.65, 0.65}
\definecolor{2nd}{rgb}{1, 0.9, 0.9}

\newcolumntype{L}{@{}l}
\newcolumntype{C}{c@{}}
\newcolumntype{R}{r@{}}

\usepackage{amssymb}%
\usepackage{pifont}%

\renewcommand{\paragraph}[1]{\vspace{.65em}\noindent\textbf{#1}.}

\definecolor{color1}{rgb}{0.9, 0.65, 0.65}
\definecolor{color2}{rgb}{0.95, 0.8, 0.8}
\definecolor{color3}{rgb}{1.0, 0.9, 0.9}

\renewcommand{\ws}[1]{#1}
\renewcommand{\cl}[1]{#1}
\renewcommand{\at}[1]{#1}

\definecolor{cvprblue}{rgb}{0.21,0.49,0.74}
\usepackage[pagebackref,breaklinks,colorlinks,citecolor=cvprblue]{hyperref}
\usepackage{orcidlink}

\usepackage{comment}

\usepackage{wrapfig}

\makeatletter
\newcommand{\printfnsymbol}[1]{%
        \textsuperscript{\@fnsymbol{#1}}%
}
\makeatother

\def\paperID{161}
\def\confName{3DV\xspace}
\def\confYear{2025\xspace}

\begin{document}
\title{{NoKSR: Kernel-Free Neural Surface Reconstruction \\ via Point Cloud Serialization}}
\author{Zhen (Colin) Li$^{1}$\thanks{Equal contribution.} \quad 
Weiwei Sun$^{2,3}$\footnotemark[1] \thanks{Work performed while at the University of British Columbia. } \quad
Shrisudhan Govindarajan$^{1}$ \quad \\ 
Shaobo Xia$^{4}$ \quad
Daniel Rebain$^{2}$ \quad
Kwang Moo Yi$^{2}$ \quad
Andrea Tagliasacchi$^{1, 5, 6}$
\\[.5em]
\small
$^{1}$Simon Fraser University \hspace{1pt}
$^{2}$University of British Columbia \hspace{1pt}
$^{3}$Amazon \hspace{1pt}
\\
\small
$^{4}$Changsha University of Science and Technology \hspace{1pt}
$^{5}$University of Toronto \hspace{1pt}
$^{6}$Google DeepMind \hspace{1pt}
\\
\texttt{\href{https://theialab.github.io/noksr/}{theialab.github.io/noksr}}
}

\maketitle

\begin{abstract}
We present a novel approach to large-scale point cloud surface reconstruction by developing an efficient framework that converts an irregular point cloud into a signed distance field~(SDF).
Our backbone builds upon recent transformer-based architectures~(i.e.~PointTransformerV3), that serializes the point cloud into a locality-preserving sequence of tokens.
We efficiently predict the SDF value at a point by aggregating nearby tokens, where fast approximate neighbors can be retrieved thanks to the serialization.
We serialize the point cloud at different levels/scales, and non-linearly aggregate a feature to predict the SDF value.
We show that aggregating across multiple scales is critical to overcome the approximations introduced by the serialization (i.e. false negatives in the neighborhood).
Our frameworks sets the new state-of-the-art in terms of accuracy and efficiency~(better or similar performance with half the latency of the best prior method, coupled with a simpler implementation), particularly on outdoor datasets where sparse-grid methods have shown limited performance.
\end{abstract}

\section{Introduction}
\label{sec:intro}

Reconstructing the surface sampled by a point cloud is a fundamental problem with many applications in robotics~\cite{tong2023scene}, autonomous driving~\cite{Autodriving}, and virtual reality~\cite{zhuang2024survey,guo2024fast}.
We tackle this task by predicting the \textit{signed distance field}~(SDF) associated with a given point cloud: a function that returns the signed distance to the nearest surface for any given 3D position.
Given the distance field, the surface can be extracted by finding the zero-crossings of the distance function.

State-of-the-art approaches such as NKSR~\cite{huang2023neural} and NeuralUDF~\cite{neuraludf} train a point cloud backbone to predict the distance value for any position in space.
Their backbones are trained on a collection of scenes, so as to capture the priors within the data that allows reconstruction to be performed even when the problem is \textit{ill-posed} (e.g., when the point cloud is sparse and/or incomplete).

\begin{figure}[t]
\centering
\includegraphics[width=\linewidth, trim= 0 140 0 0, clip]{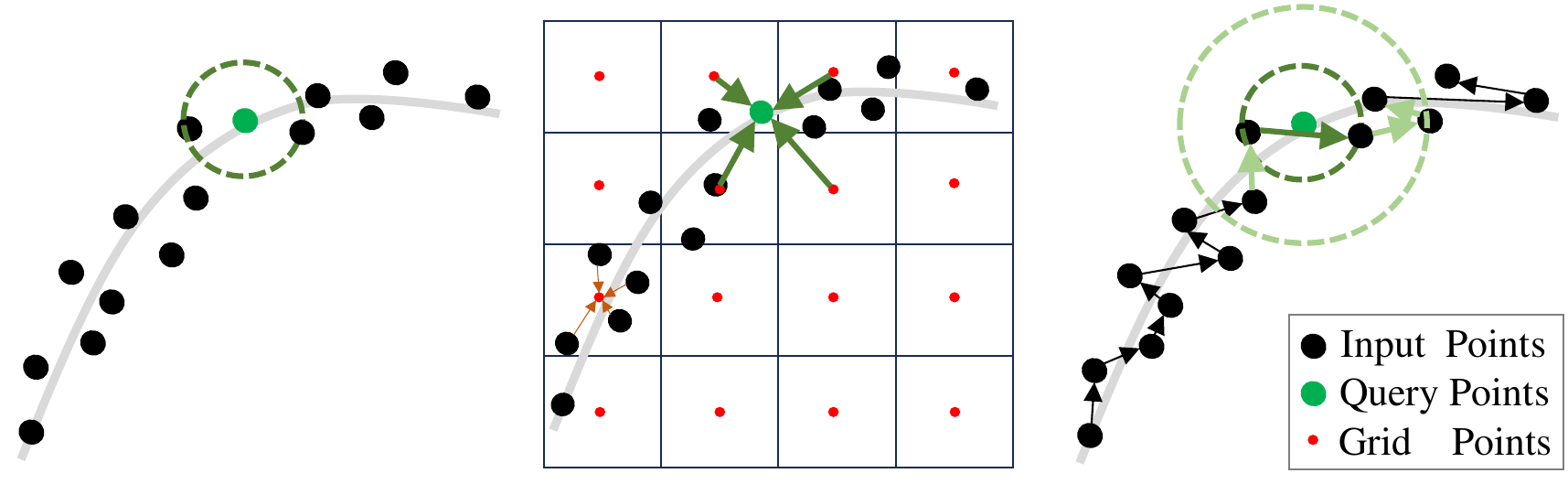}\\
\vspace{1em}
\resizebox{\linewidth}{!}{
\setlength{\tabcolsep}{4pt}

\begin{tabular}{@{}llccc@{}}
    \toprule
    & & CD ($10^{-2}$) $\downarrow$ & F-Score $\uparrow$ & Latency (s) $\downarrow$ \\
    \midrule
    K-Nearest Neighbors & oracle & 3.2 & 97.5 & 3.2 \\
    Ours~(Minkowski)~\cite{choy20194d} & grid-based & 3.8 & 96.2 & 1.5 \\
    \bf Ours & serialization & 3.3 &  97.4 & 1.7 \\
    \bottomrule
\end{tabular}

}
\caption{
To locally predict the SDF value that (implicitly) reconstructs the surface, the pivotal operation is to \textit{aggregate} the information (i.e.~features) of nearby points.
(left) Working on the point cloud directly is difficult, as there is no simple way to implement multi-scale architectures suitable for large scale point cloud processing.
(middle)~State-of-the-art methods therefore opt to quantize the input point cloud to a voxel grid, and employ established sparse CNN backbones, but quantization leads to information loss.
(right) By fetching approximate neighbors via serialization we can fetch the local context efficiently \textit{and} avoid information loss.
{We summarize the performance of representative works on a large scale outdoor dataset (\carla), and show that our method achieves the best performance in both time efficiency (latency) and accuracy (CD and F-score); for additional details see~\Cref{sec:results}. }
}
\label{fig:teaser}
\vspace{-1em}
\end{figure}

The core operation within these backbones is to predict a feature that \textit{aggregates} the information of input points near the query, which is then decoded to an SDF value.
In state-of-the-art models, these aggregation operations are realized by implementing multi-scale sparse convolutional neural networks~\cite{wang2023octformer,choy20194d}.
To be able to scale to large-scale point clouds, these backbones require voxelizing the input point cloud, and summarizing the information of points therein: a spatial \textit{quantization} operation that inevitably leads to information loss.
This quantization operation is detrimental when real-world point clouds are used, as the non-uniformity of sampling leads to performance degradation.

Rather than relying on spatial quantization and sparse-CNNs, we build upon PointTransformerV3~\cite{wu2024point}, and aggregate information by relying on locality-preserving serialization: we serialize the input point cloud to an ordered list, so that nearby points in the list are in close Euclidean proximity; see \Cref{fig:teaser}.
The serialization transformation \textit{does not} incur information loss due to quantization, and it offers superior computational efficiency in terms of feature aggregation compared to methods based on voxelization.

With serialization, retrieving the local neighbors to aggregate our features can result in \textit{false negatives}: points can be close in Euclidean space, but far in their serialized index.
To circumvent this problem, we retrieve features from approximate nearest neighbors \textit{across} several serialization levels, as provided by~\cite{wu2024point}, and then aggregate them with a PointNet architecture~\cite{PointNet} to predict the signed distance function.

Compared to state-of-the-art techniques, our framework requires neither \at{heavily engineered sparse processing backbones}~\cite{neuraludf}, nor differentiating through linear systems of equations~\cite{williams2022neural, huang2023neural}.
Nonetheless, this simple framework outperforms the state-of-the-art in \textit{both} time efficiency and reconstruction quality on \textit{multiple datasets} including ScanNet~\cite{dai2017scannet}, SceneNN~\cite{hua2016scenenn}, Carla~\cite{huang2023neural, dosovitskiy2017carla}, SyntheticRoom~\cite{peng2020convoccnet}.

Given the dominance of voxel-based data structures in surface reconstruction from point clouds, we demonstrate that carefully designed \textit{point-based} architectures can also be highly effective for this task, and we hope this will inspire \textit{renewed} interest in the research area.

\section{Related works}

Explicit 3D representations such as points~\cite{achlioptas2018learning}, voxels~\cite{tatarchenko2017octree}, meshes~\cite{fu2024lfs}, and polygonal surfaces~\cite{Nan_2017_ICCV} are commonly used for visualization and reconstruction~\cite{kazhdan2013screened}.
However, the discrete structures of these representations are challenging to adapt for learning-based approaches~\cite{takikawa2021neural,peng2020convoccnet} which rely on differentiability.
As a result, implicit 3D representations via neural networks have gained popularity ~\cite{park2019deepsdf} as they can be converted to an explicit model after training by techniques like Marching Cubes~\cite{lorensen1998marching}.
This paper addresses the task of reconstructing neural implicit surface representations from point clouds.

\paragraph{Neural implicit surface reconstruction}
Neural implicit methods utilize neural networks to model an occupancy field~\cite{mescheder2019occupancy, ouasfi2024unsupervised} or a distance field~\cite{park2019deepsdf, huang2022neural} for surface reconstruction.
Distance fields, including signed distance fields~(SDF)~\cite{park2019deepsdf,koneputugodage2024small,BaoruiTowards} and unsigned distance fields~(UDF)~\cite{chibane2020ndf,wang2022rangeudf}, are functions whose zero level set implicitly defines the object surface.
A learnt SDF predicts a query point's signed distance to the nearest surface, with a negative value indicating the point is inside the surface and a positive value indicating it is outside~\cite{takikawa2021neural}.
\quad
To encode unstructured point clouds into neural fields, various network architectures have been proposed, such as  
\begin{figure*}
  \centering
  \includegraphics[width=\linewidth]{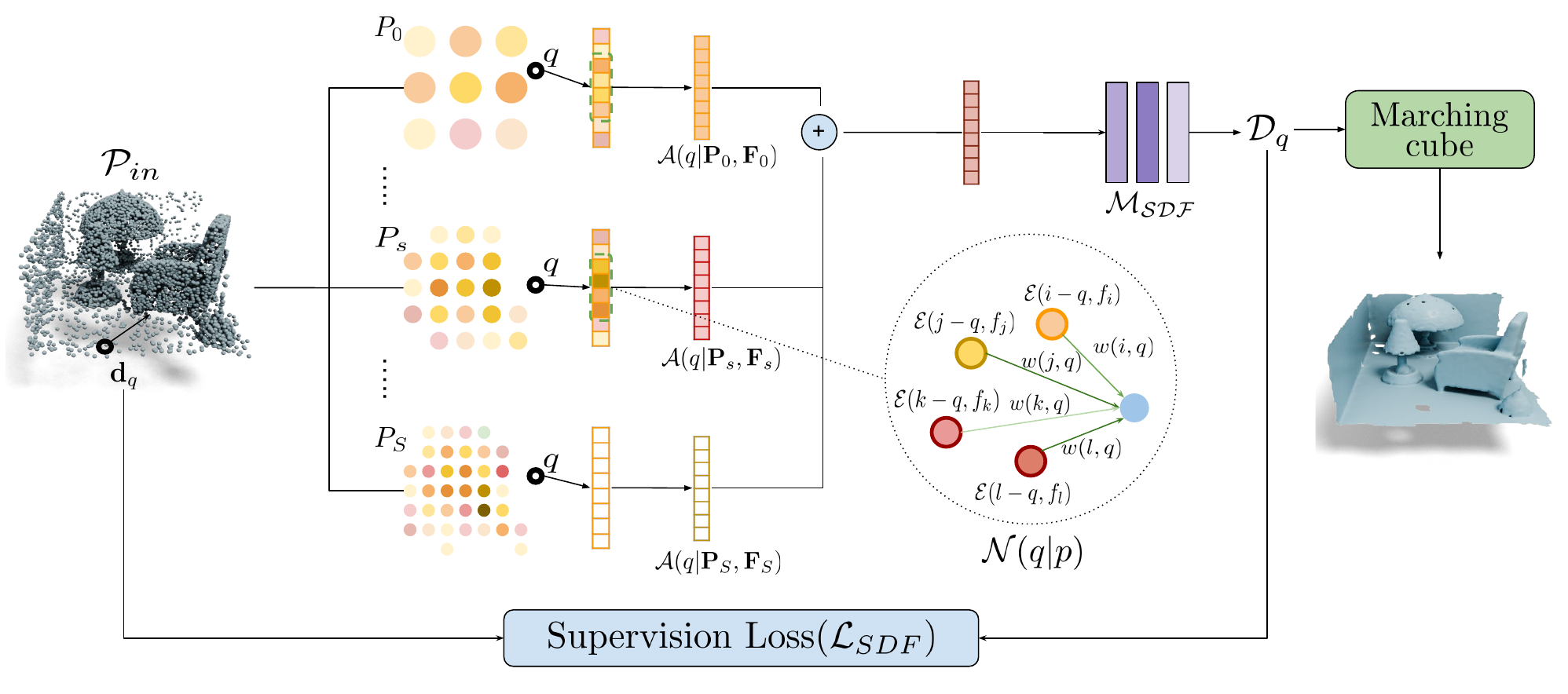}
\caption{
{{\bf Overview}}: We map a sparse input point cloud with a point cloud backbone~\cite{wu2024point} into a point feature hierarchy, from which we compute the signed distance of a query.
At each level, we utilize the efficient procedure defined in~\Cref{sec:neighbor_func} to retrieve local neighborhoods of the query. 
We then compute per-level features with the aggregation module defined in~\Cref{sec:aggregation}.
At last, we sum per-level features and convert it into the signed distance with an MLP. 
}
\label{fig:overview}
\end{figure*}

MLPs~\cite{Chen_2019_CVPR,mescheder2019occupancy,park2019deepsdf}, infinitely-wide-ReLU networks~\cite{williams2021neural}, PointNet~\cite{williams2022neural,tang2021SACon}, 3D-UNet~\cite{wang2023alto}, RandLA-Net~\cite{wang2022rangeudf}, sparse hierarchical networks~\cite{huang2023neural}, and MinkowskiNet~\cite{neuraludf}.
Compared to encoding the point cloud into a global feature~\cite{park2019deepsdf,Points2Surf}, organizing point clouds into regular or irregular grids or voxels for feature learning and spatial querying preserves more details~\cite{li2022learning,zhang20223dilg,zhong20243d,peng2020convoccnet,williams2022neural,li2024gridformer}.
For example, \cite{takikawa2021neural} uses a voxel octree to collect point-wise features, and retrieves the query point feature via trilinear interpolation at each tree level.
Alternatively, feature interpolation can be implemented with an attention-based~\cite{wang2023alto} or learning-based approach~\cite{boulch2022poco}.
Such data-driven methods often struggle to guarantee the accuracy of learned surfaces, and are difficult to scale and generalize~\cite{huang2022neural,williams2022neural}.
\citet{huang2023neural} addresses this problem by solving complex kernel functions on hierarchical voxels.
However, the quantization inherent to voxels or grids leads to information loss, and the solver increases the reconstruction time quadratically with the number of grid cells.
To solve these problems, we propose a point-based framework powered by a serialization encoding for implicit surface reconstruction.

\paragraph{Efficient point cloud networks}
Point-based networks~\cite{PointNet,hu2019randla} \at{achieve good performance on small datasets, but in applications to large point cloud data their message-passing strategy is not sufficiently computationally efficient.}
Sparse convolution networks~\cite{choy20194d,graham20183d} based on voxelization are fast but suffer from information loss. 
The recently proposed OctFormer~\cite{wang2023octformer} and PointTransformerV3~\cite{wu2024point} provide a superior combination of efficiency and encoding performance by leveraging a serialization-based strategy, and our method builds upon these approaches.

\paragraph{Point cloud serialization}
Backbone networks that rely on voxelization (e.g., MinkowskiNet~\cite{choy20194d} and sub-manifold sparse U-Net~\cite{graham20183d}) suffer of high computational cost, and from information loss due to quantization.
To avoid these limitations, point cloud serialization methods encode irregular point clouds into sequential structures with the use of space-filling curves.
This bijective encoding scheme excels at dimension reduction, preserving topology and retaining locality, making it a promising approach to address voxelization issues~\cite{wang2005space, wang2017cnn}.
\citet{chen2022efficient} leverages the Hilbert curve~\cite{hilbert1935stetige} to map voxels into an ordered sequence, enabling the use of 2D convolution and Transformers on 3D voxels.
\citet{wang2023octformer}, by employing z-order curves~\cite{morton1966computer} to sort octree nodes, achieves equal-sized point partitions and constructs an effective octree attention module for point clouds.
The idea of equal-sized sorting of windows is also adopted in~\cite{liu2023flatformer}.
To mitigate the computational overhead of K-Nearest Neighbor (KNN), \citet{wu2024point} integrates z-order and Hilbert curves to map 3D points into structured sequences and patches, upon which attention layers are constructed.
\citet{zhang2024voxel} introduces a Hilbert input layer for serializing 3D voxels, laying the foundation for a voxel-based state space model designed for 3D object detection.
This approach eliminates the need for 1D sequence grouping and padding.
\quad 
In our work, we present a point feature retrieval algorithm that operates directly on points based on point cloud serialization, and demonstrate its performance for neural surface reconstruction applications.

\section{Method}
\label{sec:formatting}
The overview of our method is illustrated in~\Cref{fig:overview}. 
Given a point cloud $\points {\in} \real^{N\times 3}$ of N points in 3-dimensional space, we compute hierarchical point features $\features$ with $S$ levels using a point-based transformer $\backbone$ parameterized via $\param_{\backbone}$ as
\begin{align}
\{(\points_s, \features_s)\}^{S}_{s=1} = \backbone(\points; \param_{\backbone})
\end{align}
where with the $s$ subscript we denote the point cloud and learned features at $s$-th level.  
For a given query $\query \in \real^{3}$, we employ the features from the feature hierarchy to predict its distance field value $\distance$:
\begin{align}
\distance = \distancefield(\query ~|~ \{(\points_s, \features_s )\} ; \param_\distancefield)
\end{align}
where $\distancefield$ has learnable parameters~$\param_{\distancefield}$.

\subsection{Distance field\texorpdfstring{ -- $\distancefield$}{}}
For each query $\query$, we calculate a per-level feature from the feature hierarchy through an \textit{aggregation} module $\aggregation$, and then sum the features of all levels as the query's feature, which is then mapped to an SDF value by $\mlp$:
\begin{align}
\distancefield(\query) = \mlp\left(\sum_{s=1}^{S} \aggregation\left(\query| \points_s, \features_s\right) \right)
\end{align}
Following~\citet{huang2023neural}, $\mlp$ is simply an MLP with a single hidden layer followed by a \textit{tanh} activation, and its parameters are included in the set~$\param_\distancefield$. 
We now describe the aggregation module $\aggregation$ in more details.

\subsection{Aggregation module -- \texorpdfstring{$\aggregation$}{}}
\label{sec:aggregation}
At the $s$-th level, we retrieve the local neighborhood at the query location and use a PointNet-style network to map the local point cloud into the per-level feature:
\begin{align}    
\aggregation(\query | \points_s, \features_s) = 
\frac{
\sum_{\point \in \neighbor(\query | \points_s)} w(\point, \query) \cdot \pointnet(\point - \query, \feature_\point)  
}{
\epsilon + \sum_{\point \in \neighbor(\query | \points_s)} w(\point, \query) 
}
\end{align}
where  $w(\point, \query)$ is the inverse spatial distance, 
$\epsilon{=}1e{-}8$ avoids division by zero, $\pointnet$ is a small MLP whose parameters are included in the set $\param_\distancefield$, $\feature_\point$ is the feature in $\features$ at $\point$, $\neighbor(\query | \point)$ is a function that retrieves the local neighborhood of $\query$ from $\points_s$. 

\begin{figure}
    \centering
    \includegraphics[width=\linewidth]{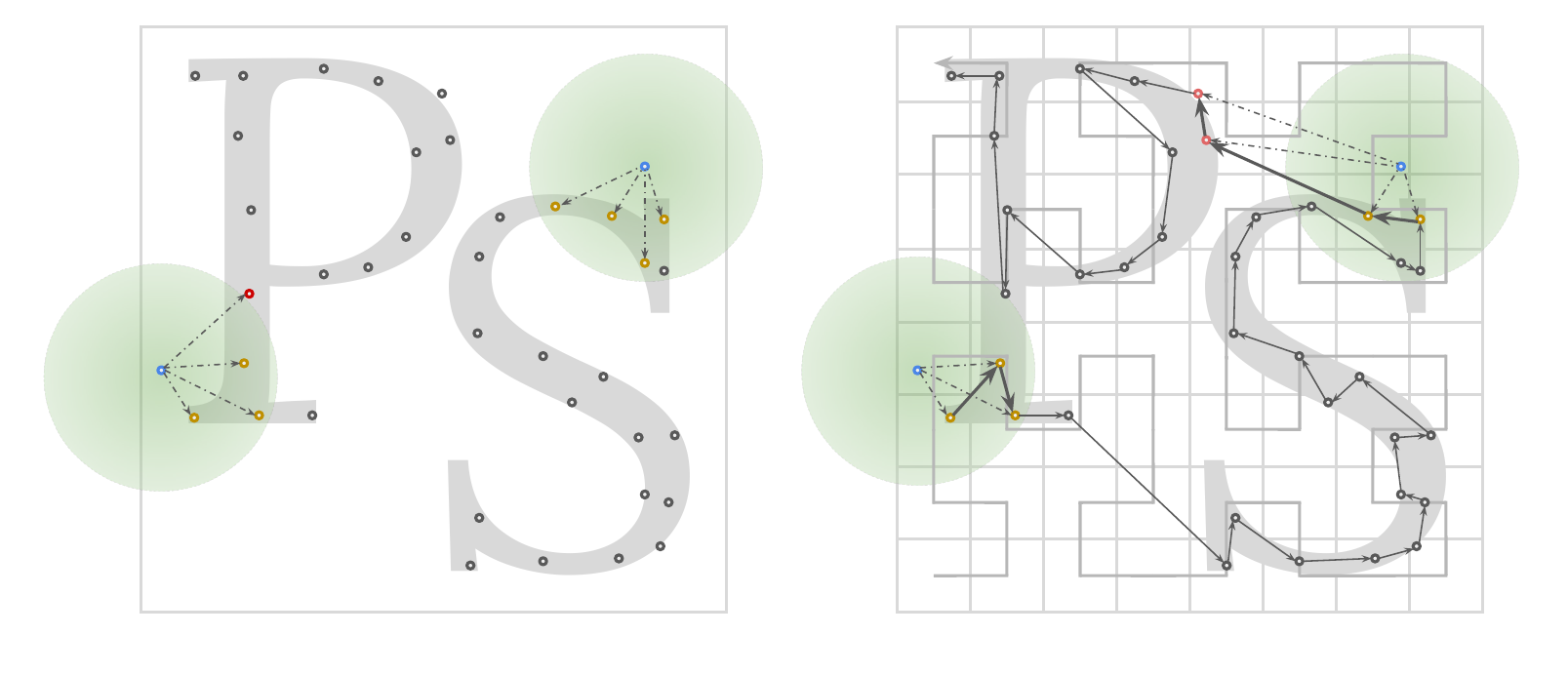}
    \caption{
    \textbf{Neighborhood function} -- (left) retrieving a local neighborhood with K-nearest neighbor(KNN) or ball-query methods is challenging to implement efficiently on GPU hardware.
    (right) we propose to retrieve a neighborhood from a 1-D ordered list, by serializing points along a Hibert curve~\cite{hilbert1935stetige}, and excluding the impact of points distant from the query~(i.e. remove false positives).
    }
    \label{fig:neighbor_func}
\vspace{-1em}
\end{figure}

\subsubsection{Neighborhood function \texorpdfstring{ -- $\neighbor$}{}}
\label{sec:neighbor_func}
Retrieving neighbors via k-nearest neighbor (KNN) or ball-query methods would be optimal, but these are difficult to implement efficiently on GPU hardware.
As the reconstruction pipeline is sensitive to the computational cost of this operation, we choose to leverage a more efficient strategy.
In particular, we implement our approximate neighborhood lookup $\neighbor$ on the locality-preserving \textit{serialization} encoding proposed by~\cite{wang2023octformer, wu2024point}. 
A serialization encoding is a hash function ($\gamma: \real^3 \hookrightarrow \mathbb{Z} $) that maps a point to a integer.
Given a point $point \in \real^{3}$, we calculate the integer as 
\begin{align} 
\gamma = \phi(\lfloor{ p/g }\rfloor))
\end{align}
where $\lfloor{ p/g }\rfloor$ is a floor function that quantizes a point with real-valued coordinates to the integral coordinates of cells in a 3-dimensional grid with size $g$, and $\phi$ is a bijective function that maps 3D coordinates $\mathbb{Z}^3$ to 1D values $\mathbb{Z}$. 
We define the bijective $\phi$ as a space filling curve, which traverses 3D space in a \textit{locality-preserving} order. 
We utilize Hilbert curves~\cite{hilbert1935stetige}, and to avoid collisions in the quantization of point coordinates to grid cells, we use a very fine grid resolution across all levels, as there is no cost associated with increasing the resolution of this \textit{virtual} grid.
As illustrated in \Cref{fig:neighbor_func}, to retrieve the local neighborhood of a query from a point cloud, we first encode the point cloud into a set of sorted integers using $\gamma$.
We then apply $\gamma$ to the query coordinate and search through its neighbors on the 1D line to identify close-by points in Euclidean space.

\subsection{Training}
To train our networks, we optimize the loss:
\begin{align}    
\argmin_{\param_\backbone, \param_\distancefield, \param_{\mathcal{C}}}  \:\: \lambda_{\text{SDF}} \loss{SDF} + \lambda_{\text{Eikonal}} \loss{Eikonal} + \lambda_{\text{mask}} \loss{mask}
\end{align}
where $\lambda_{\text{SDF}}$ , $\lambda_{\text{eikonal}}$ and $\lambda_{\text{mask}}$ are the coefficients for loss terms, which we will detail below. 

\paragraph{Signed distance function supervision -- $\loss{SDF}$}
We define $\loss{SDF}$ to reproduce the ground truth SDF value $\distance_q$ at $\query$:
\begin{align*}
\loss{SDF} = \expect_{\query\sim\mathcal{Q}} 
\:
[|| \distance_\query - \distancefield(\query) ||_1]    
\end{align*}
where $\mathcal{Q}$ is the distribution from~\citet{huang2023neural}.

\paragraph{Surface regularizer -- $\loss{Eikonal}$}
We regularize the field $\distancefield$ with an Eikonal loss to encourage this function to be a signed distance field away from the surface: 
\begin{align}
    \loss{Eikonal} = \expect_{\x \sim \mathcal{Q}} [(||\nabla_\x \distancefield(\x)\|_2 - 1)^2 ]
\end{align}

\paragraph{Auxiliary loss -- $\loss{mask}$}  
\at{Following~\cite{huang2023neural}, the classification branch $\mathcal{C}$ with learnable parameters $\param_{\mathcal{C}}$ classifies queries as near/far from the surface as supervised by the loss:}
\begin{align*}
\loss{mask} = \expect_{\query\sim\mathcal{Q}} [\textbf{CE}(\mathbf{c}_\query, \mathcal{C}(\query)
)] 
\end{align*}
where \textbf{CE} is the cross entropy function, $\mathbf{c}_\query$ is ground-truth binary label calculated by thresholding the ground-truth SDF with empirically chosen values of 0.015 meters for indoor scenes, and 0.1 meters for outdoor.
\at{At inference time, the output of this classifier helps avoid reconstructing surfaces that are far from the input point cloud~(i.e. not supported by input point-cloud data).}

\section{Results}
\label{sec:results}
Following \nksr, we evaluate our method using metrics including the standard Chamfer-$L_1$ Distance~(CD-$L_1 \times 10^{-2}$, $\downarrow$) and F-score~($\uparrow$) with a threshold~($\delta{=}0.010$). 
We also report additional metrics proposed in \nksr~including Chamfer-$L_1$ Distance by Completeness (Comp.~$\times 10^{-2}$, $\downarrow$) and Accuracy (Acc.~$\times 10^{-2}$, $\downarrow$) in the \texttt{Supplementary Material}. 
We evaluate our method on multiple datasets, under two settings including in-domain evaluation for accuracy estimation -- training set and test set are from same dataset, and cross-domain evaluation for generalization ability estimation where training set and test set are from different datasets. 
Additionally, for cross-domain evaluation we use the following datasets prepared by the leading voxel-based baseline, \nksr, and one additional dataset from RangeUDF~\cite{wang2022rangeudf}:

\begin{itemize}
    \item \synthetic{}  is a synthetic dataset created from ShapeNet objects~\cite{chang2015shapenet}. Each scene contains 2-3 objects. 
    Following prior works~\cite{wang2022rangeudf,chibane2020ndf}, we re-scale the synthetic rooms to roughly match real-world scale.
    There are 3750 scenes as training set and \ws{995 scenes} as the test set. 
    \item \scannet{} is a real-world indoor scene dataset. We use the setting from previous work~\cite{wang2022rangeudf, tang2021SACon, peng2020convoccnet, boulch2022poco} where we train on 1201 rooms and test on 312 rooms. 
    \item \carla is a large-scale outdoor driving scene prepared by NKSR~\cite{huang2023neural} using the CARLA simulator~\cite{dosovitskiy2017carla}. 
    \ws{Following NSKR~\cite{huang2023neural}, we test on two subsets including the 'Original' subset (10 random drives simulated on 3 towns) and the 'Novel' subset (3 drives from an additional town only for testing).}
    To avoid exploding GPU memory during training, we follow NKSR~\cite{huang2023neural} to divide a large scene into patches. The resultant training set has {3757} patches. 
    \item \scenenn{}  is a real-world indoor dataset prepared by RangeUDF~\cite{wang2022rangeudf} which we used for cross-domain evaluation. We only use its test set which consists of 20 scenes.
\end{itemize}

\begin{table*}
\centering
\resizebox{\linewidth}{!}{
\setlength{\tabcolsep}{3pt}
\begin{tabular}{LccccccccccccC}
\toprule
Methods & & \multicolumn{3}{c}{\ws{{\bf \synthetic}}}  &  \multicolumn{3}{c}{{\bf \scannet}} & \multicolumn{3}{c}{\ws{{\bf \carla(Original)}}} & \multicolumn{3}{c}{\ws{{\bf \carla(Novel)}}} \\ 
 \cmidrule(lr){3-5} \cmidrule(lr){6-8} \cmidrule(lr){9-11} \cmidrule(lr){12-14} 
&Primitive& CD ($10^{-2}$) $\downarrow$ & F-Score  $\uparrow$ & Latency (s) $\downarrow$  & CD ($10^{-2}$) $\downarrow$ & F-Score  $\uparrow$ & Latency (s) $\downarrow$  & CD (cm) $\downarrow$ & F-Score  $\uparrow$ & Latency (s) $\downarrow$ & CD (cm) $\downarrow$ & F-Score  $\uparrow$ & Latency (s) $\downarrow$ \\        
\midrule
SA-CONet~\cite{tang2021SACon} & Voxels & {0.496} & {93.60} & - & - & - & - & - & - & - & - & - & -\\
ConvOcc~\cite{peng2020convoccnet} & Voxels & {0.420} & {96.40} & - & - & - & - & - & - & - & - & - & -\\
NDF~\cite{chibane2020ndf} & Voxels & {0.408} & {95.20} & - & 0.385  & 96.40  & -  & - & - & - & - & - & -\\
RangeUDF~\cite{wang2022rangeudf} & Voxels & {0.348} & {97.80} & {-} & 0.286 & 98.80 & - & - & - & - & - & - & -\\
\ws{TSDF-Fusion~\cite{zeng20163dmatch}} & -  & - & - & - & - & - & - & 8.1 & 80.2 & - & 7.6 & 80.7 & - \\
\ws{POCO~\cite{boulch2022poco}} & - & - & - & - & - & - & - & 7.0 & 90.1 & - & 12.0 & 92.4 & - \\
\ws{SPSR~\cite{kazhdan2013screened}} & - & - & - & - & - & - & - & 13.3 & 86.5 & - & 11.3 & 88.3 & - \\
\nksr & Voxels &  \underline{0.346} &  \underline{97.41} & \underline{0.40} & \underline{0.246} & \underline{99.51} & \underline{1.54} &  \underline{3.9} &  \underline{93.9} &  \underline{2.0} &  \underline{2.9} &  \underline{96.0} &  \underline{1.8} \\
\nksr (more data) & Voxels & - & - & - & - & - & - & {3.6} & {94.0} & {2.0} & {3.0} & {96.0} & {1.8}\\
Ours~(Minkowski)~\cite{choy20194d} \scriptsize{(w/ KNN)} & Voxels & - & \todo{} & \todo{} & 0.254 & 99.41 & 0.46 & 3.4 & 97.2 &1.9 & 2.7 & 98.1 & 2.0 \\
Ours~(Minkowski)~\cite{choy20194d} & Voxels & - & \todo{} & \todo{} & 0.301 & 98.48 & 0.31 & 3.8 & 96.2 & 1.5 & 3.0 & 97.4 & 1.5\\
\rowcolor{1st} Ours \scriptsize{(w/ KNN)} & Points &{0.321} & {98.34} & {0.13} & {0.243} & {99.61} & {0.48} &{3.2} & {97.5} & {3.2} &{2.6} & {98.3} & {3.4}\\
\rowcolor{1st}Ours & Points & {0.360} & {96.32} & 0.14 & 0.257 & 99.33 & 0.49 & {3.3} & {97.4} & 1.7 & {2.7} & {98.2} & 1.7 \\

\bottomrule
\end{tabular}
}
\caption{\textbf{In-domain evaluation} -- We show that our method achieves the best accuracy (CD and F-score) with significantly improved time efficiency~(inference latency).
Note we retrain \nksr (numbers are underlined) for fairer comparison, \ws{as the training data for \nksr is different from ours -- i.e., they reported some models trained on a ``mix'' of datasets, which is impossible to reproduce.
}
}
\label{tab:indomain}
\end{table*}

\paragraph{Evaluation pipeline}
To evaluate our method, we first extract the mesh with Dual Marching Cubes~\cite{schaefer2004dual} on the predicted SDF, and then compute the CD and F-score between 100k points sampled on the mesh, and 100k points sampled from the ground-truth dense point cloud.
We use the same approach as \nksr to prepare the input point clouds for training and evaluation from the ground-truth dense point clouds through downsampling.
Specifically, for indoor datasets (i.e., \synthetic, 
\scannet and \scenenn), we uniformly sample 10K points sampled from the ground truth dense point cloud. 
For outdoor driving scenes~(i.e., \carla), we follow the evaluation pipeline from \nksr.
We sample sparse input point clouds with a sparse 32-beam LiDAR with a ray distance noise of 0-5 cm and pose noise of $0-3^\circ$, and obtain the ground truth from a noise-free dense 256-beam LiDAR.

\begin{figure*}
\centering
\includegraphics[width=\linewidth]{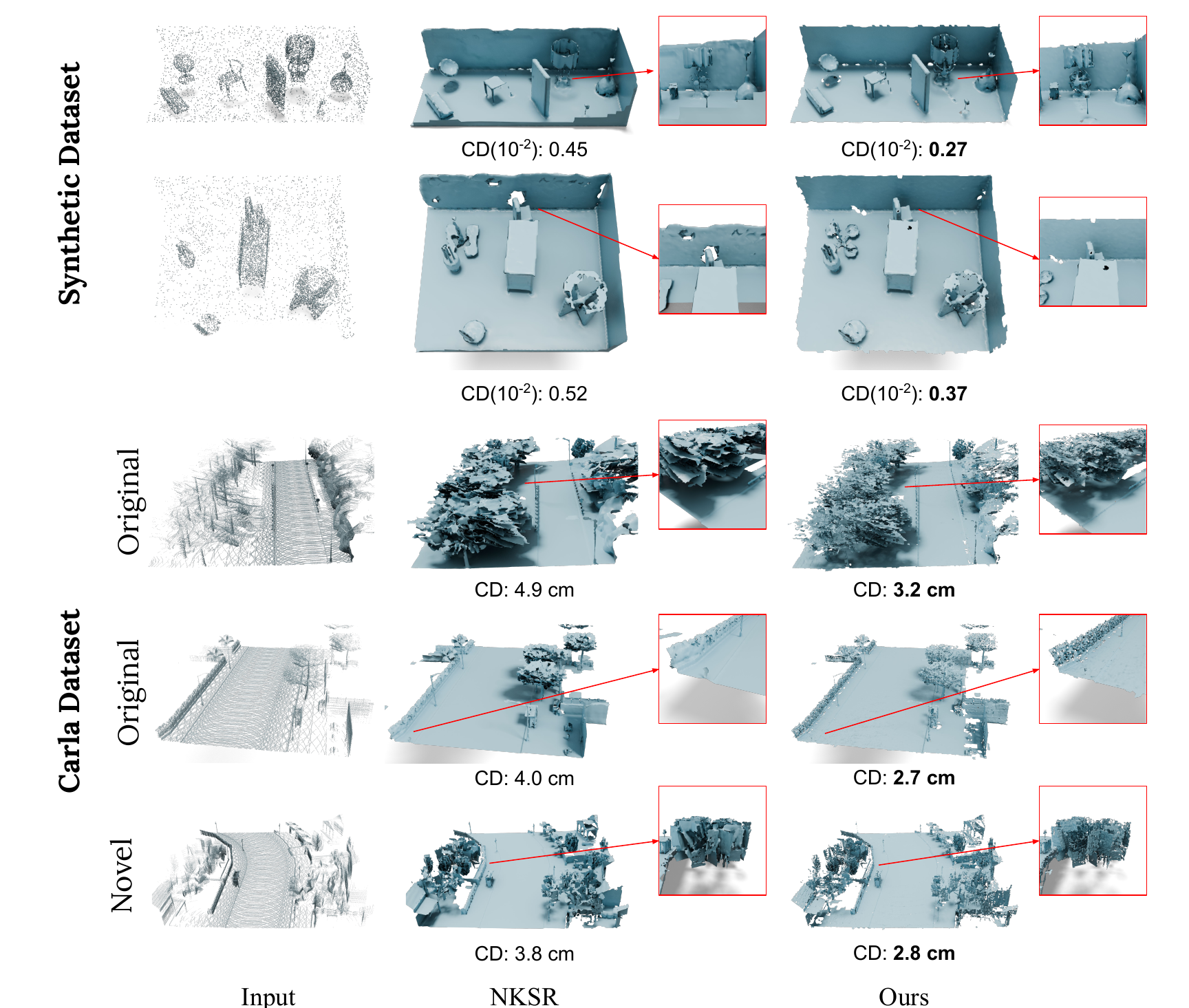}
\caption{
{\textbf{Qualitative results on \carla and \synthetic}} -- our method achieves high quality surface reconstructions which preserve more details than \nksr~which loses information due to quantization for large and non-uniformly sampled datasets like Carla.
}
\label{fig:qual_results_carla_syn}
\end{figure*}
 
\begin{figure*}
\centering
\vspace{-1em}
\includegraphics[width=.95\linewidth]{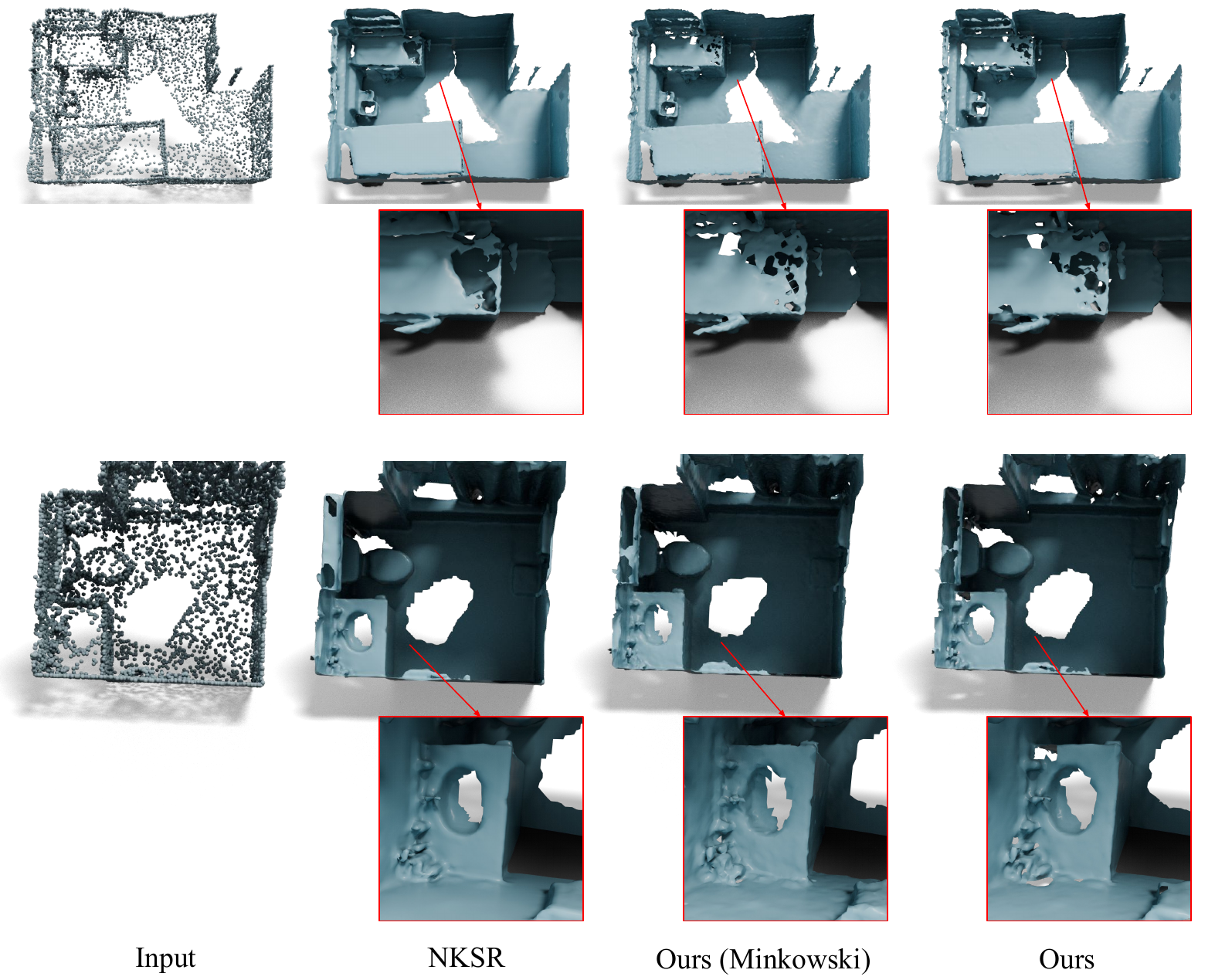}
\caption{
Qualitative results on \scannet: We compare our method with prior SOTA~\cite{huang2023neural} and Ours~(Minkowski)~\cite{choy20194d} that is more comparable as it only differs from ours in the backbone. Our method achieves reconstruction of similar quality to the SOTA. It also \textit{significantly} outperforms Ours~(Minkowski), highlighting the importance of point-based methods. 
}
\vspace{-1em}
\label{fig:scannet_results}
\end{figure*}

\paragraph{Implementation details}
We base our feature backbone on PointTransformerV3~\cite{wu2024point} with 4-levels.
The PointNet-style network is a 2-layered residual connection MLP, with hidden dimension of $32$ and output feature dimension of $32$.    
The grid size used in neighborhood function is $0.01$ meters.
Following \nksr, we use the similar coefficients for loss terms -- i.e., $\lambda_{\text{SDF}}$ is $300$ and $\lambda_{\text{mask}}$ is $150$.
However, we empirically set $\lambda_{\text{Eikonal}}$ to $10$~(\nksr does not need this regularizer thanks to its specialized surface solver).
We train our model with a batch size of $4$ on either a single \texttt{NVIDIA RTX A6000 ADA} or an \texttt{NVIDIA L40S}, and a learning rate of $10^{-3}$.
We adopt the Adam optimizer with default parameters.
We set the maximum number of epochs to 200 and employ a cosine learning rate decay starting from epoch 120.

\begin{table*}
\centering
\resizebox{\linewidth}{!}{
\setlength{\tabcolsep}{2pt}
\begin{tabular}{LccccccccccC}
\toprule
Methods & & \multicolumn{3}{c}{{\bf \synthetic $\rightarrow$ \scannet}}  &  \multicolumn{3}{c}{{{\bf \scannet $\rightarrow$ \synthetic}}} & \multicolumn{3}{c}{{{\bf \scannet $\rightarrow$ \scenenn}}} \\ 
 \cmidrule(lr){3-5} \cmidrule(lr){6-8} \cmidrule(lr){9-11}
&Primitive& CD ($10^{-2}$) $\downarrow$ & F-Score  $\uparrow$ & {Latency (s) $\downarrow$ } & CD ($10^{-2}$) $\downarrow$ & F-Score  $\uparrow$ & {Latency (s) $\downarrow$ } & CD ($10^{-2}$) $\downarrow$ & F-Score  $\uparrow$ & {Latency (s) }$\downarrow$ \\       
\midrule
SA-CONet~\cite{tang2021SACon} & Voxels & 0.845 & 77.80 & - & - & - & - & - & - & - \\
ConvOcc~\cite{peng2020convoccnet} & Voxels & 0.776 & 83.30  & - & - & - & - & - & - & - \\
NDF~\cite{chibane2020ndf} & Voxels & 0.452 & 96.00 & - & {0.568} & {88.10} & - & 0.425 & 94.80 & - \\
RangeUDF~\cite{wang2022rangeudf} & Voxels & {0.303} & {98.60} & {-} & 0.481& 91.50 & - & 0.324 & 97.80 & - \\
\nksr & Voxels & {0.329} & {97.37} & {2.02} & {0.351} & {97.41} & {0.46} & {0.268} & {99.18} & {1.95} \\
\rowcolor{1st} Ours (w/ KNN) & Points & {0.284} & {98.65} & {0.54} & {0.327} &{98.37} & {0.13} & {0.277} & {99.00} & {0.50} \\
\bottomrule
\end{tabular}
}
\caption{\textbf{Cross-domain evaluation} -- we achieve the best generalization ability in two cases with much better time efficiency. In the other case where we generalize from \scannet to \scenenn, we achieve accuracy on par with the SOTA baseline~\cite{huang2023neural} with less than a half of their latency.  
}
\vspace{-1.4em}
\label{tab:across_domain}
\end{table*}

\paragraph{Reconstruction latency}
For both our models and NKSR, we record the reconstruction latency for all indoor scenes on a single \texttt{NVIDIA RTX 3090}, and for large outdoor scenes on a single \texttt{NVIDIA L40s} given that more GPU memory is required.
We omit data loading time, and only record the average forward pass time. 

\subsection{In-domain evaluation}
We compare against \nksr~(the current state-of-the-art), RangeUDF~\cite{wang2022rangeudf},  SPSR~\cite{kazhdan2013screened}, NDF~\cite{chibane2020ndf}, ConvOcc~\cite{peng2020convoccnet} and SA-CONet~\cite{tang2021SACon}.     
We further include a baseline that replaces our backbone with MinkowskiNet~\cite{choy20194d} (i.e., Ours~(Minkowski)) to show the degraded performance due to the information loss caused by voxelization.

\paragraph{Quantitative results -- \Cref{tab:indomain}}
Across indoor and outdoor datasets, our method outperforms baselines in terms of accuracy and time efficiency. Especially in outdoor datasets, our method achieves the best surface reconstruction with the smallest latency -- nearly \textit{half} of the second best's latency.
In indoor datasets, which have relatively uniform sampling patterns, we achieve accuracy on par with the previous state-of-the-art, but with significantly improved time efficiency.
Note that we achieve this advantage even with KNN because, in smaller indoor point clouds, the highly engineered KNN implementation has similar time efficiency to that of our neighborhood function.
We further detail our analysis on this matter in the \texttt{Supplementary Material}. 
We also note that our approximate neighborhood function is still effective, as it outperforms the directly comparable baseline MinkowskiNet~\cite{choy20194d}, which shares the same structure except for the backbone and neighborhood function.

\paragraph{Qualitative results -- \Cref{fig:qual_results_carla_syn,fig:scannet_results}}
We show that our method tends to reconstruct surfaces of the best quality among the compared methods.
Especially, on the non-uniform large scale \carla, our method tends to preserve more details than the previous state-of-the-art~\cite{huang2023neural}, which voxelizes the point cloud.   

\subsection{Cross-domain evaluation -- \Cref{tab:across_domain}}
We further test the generalization ability of our method with a cross-domain evaluation.
We evaluate models trained with dataset A on other a different dataset B; we denote this as~A $\rightarrow$ B. 
As shown in \Cref{tab:across_domain}, there are three cases in total.
In two cases (i.e., \synthetic $\rightarrow$ \scannet and \scannet $\rightarrow$ \synthetic), our method achieves the best accuracy with the best time efficiency. 
In another case (\scannet $\rightarrow$ \scenenn), we achieve accuracy on par with SOTA~\cite{huang2023neural} with a much better time efficiency, i.e., less than a half of the latency required by the SOTA~\cite{huang2023neural}.

\subsection{Ablation studies}
Our ablations are executed on \scannet, as it is a real-world dataset, and is equipped with precise ground truth surface meshes.

\begin{table}
\centering
\resizebox{.9\columnwidth}{!}{
\begin{tabular}{LccccccC}
\toprule
{\bf Neighbor Num.} & {CD (10\textsuperscript{-2})} $\downarrow$ & {F-score} $\uparrow$ & Latency (s) $\downarrow$ \\ \midrule
 2 & 0.246 & 99.56 & 109 \\
 4 & 0.244 & 99.59 & 127 \\
 \rowcolor{1st} 
8 & {0.243} & 99.61 & 151 \\
16 & 0.256 & 99.28 & 187 \\
\bottomrule
\end{tabular}
}
\caption{{\bf The impact of neighborhood size} -- larger neighborhoods lead to increased computational cost, and we find that 8 neighbors gives the best balance of cost and quality.}
\label{tab:numpts_neighbor}
\vspace{-1em}
\end{table}

\paragraph{Impact of neighborhood size -- \Cref{tab:numpts_neighbor}}
We analyze the impact of neighborhood size on performance. Larger neighborhood size leads to increased computation overhead. 
We show that the 8-nearest neighboring points gives the best trade-off between accuracy and time efficiency.
Considering a large number (e.g., 16) of neighboring points degrades performance as the the aggregation module has limited capacity to predict the precise SDF from a large local point cloud.

\begin{table}
\centering
\resizebox{.95\columnwidth}{!}{
\begin{tabular}{@{}lcccccc@{}}
\toprule
\makecell{\bf Num. of hidden\\\bf layers in $\aggregation$} & CD (10\textsuperscript{-2}) $\downarrow$ & F-score $\uparrow$ & Latency (s) $\downarrow$ \\ \midrule
 2 & 0.257 & 99.33 & 152 \\
 4 & 0.256 & 99.32 & 166 \\
\bottomrule
\end{tabular}
}
\caption{{\bf Impact of capacity of $\aggregation$} -- we find that increasing the number of layers in $\aggregation$ beyond 2 decreases time efficiency without substantially improving the reconstruction quality.}
\label{tab:agg_capacity}
\vspace{-1em}
\end{table}

\paragraph{Impact of capacity of $\aggregation$ -- \Cref{tab:agg_capacity}} 
We report how the capacity of the aggregation module $\aggregation$ (i.e., different number of hidden layers) impacts the performance.
We observe that aggregation modules of higher capacity give better performance but degraded time efficiency. However, as shown in~\Cref{tab:agg_capacity}, a very large capacity (4 layers) for $\aggregation$ does not help.
We show that we we use 2 layers to have a good trade-off between accuracy and time efficiency. 
We supplement~\Cref{tab:agg_capacity} with an analysis across even more levels in the \texttt{Supplementary Material}.

\begin{table}
\centering
\resizebox{.9\columnwidth}{!}{
\begin{tabular}{@{}lcccc@{}}
\toprule
\textbf{Num. of scales} &KNN & Minkowski & Z-order & Hilbert  \\ \midrule
0 & 1.00 & 0.17 & 0.44  & \cellcolor{1st}0.46  \\
1 & 1.00 & 0.29 & 0.48  & \cellcolor{1st}0.50  \\
2 & 1.00 & 0.38 & 0.49  & \cellcolor{1st}0.52  \\
3 & 1.00 & 0.44 & 0.49  & \cellcolor{1st}0.53  \\ %
\bottomrule
\end{tabular}
}
\caption{\textbf{Recall rate of our Hilbert-curve based $\neighbor$} -- we find that the Hilbert curve consistently outperforms both the Z-order curve~\cite{morton1966computer} and the one-ring neighborhood from Minkowski relative to the exact k-nearest neighbors.
}
\vspace{-1em}
\label{tab:locality_neighbor}
\end{table}

\paragraph{Analysis of neighbors retrieved by~$\neighbor$ -- \Cref{tab:locality_neighbor}}
\at{We now investigate the quality of the point neighborhoods retrieved by various possible implementations for $\neighbor$.
In particular, we are interested to experimentally study whether our serialization indeed preserves locality.
To quantify this, we treat the neighborhood retrieved with KNN as the ground-truth.}
We report the recall rate of a local neighborhood by comparing it with this ground truth~(we ignore the precision rate because we remove false positives with a distance threshold).
We also report the recall rate of the one-ring neighborhood retrieved in Minkowski~\cite{choy20194d}.
We show that the recall rate of our Hilbert $\neighbor$ is the best across variants, and across all scales.

\begin{table}[t]
\centering
\resizebox{\columnwidth}{!}{
\begin{tabular}{L rr rR}
\toprule
Methods & \multicolumn{2}{c}{Uniform} & \multicolumn{2}{c}{Non-Uniform}   \\ 
\cmidrule(r){1-1}
\cmidrule(lr){2-3}
\cmidrule(l){4-5}
\nksr & 0.246 & 480s & 0.273 & 668s  \\
Ours~(Minkowski)~\cite{choy20194d}  & 0.301 & 97s & 0.349 & 94s \\
Ours~(Minkowski)~\cite{choy20194d} {(w/ KNN)} & 0.254 & 145s & 0.294 & 155s \\
\rowcolor{1st} Ours~(w/ serialization) & {0.257} & {152s} & {0.296} & {145s} \\
\rowcolor{1st} Ours~(w/ KNN) & \textbf{0.243} & \textbf{151s} & \textbf{0.273} & \textbf{142s}  \\
\bottomrule
\end{tabular}
}
\caption{
\textbf{The impact of sampling} -- we evaluate uniform vs non-uniform sampling on ScanNet. We find that our method achieves the best accuracy (in terms of CD ($10^{-2}$)) and good time efficiency compared to \nksr~for both sampling types.
}
\vspace{-1em}
\label{tab:nonuniform_scannet}
\end{table}

\paragraph{The impact of sampling pattern --~\Cref{tab:nonuniform_scannet}} 
We report the impact of sampling pattern on performance by evaluating models on ScanNet point clouds that are uniformly or non-uniformly sampled. 
{To non-uniformly sample the ScanNet point clouds, we first partitioned the scene into eight blocks and randomly sampled a different number of points from each block. The number of samples followed an arithmetic sequence with a common difference of 200. Finally, we padded the last block to ensure that the total number of points remained 10K.}
 
We show that our method achieves better robustness to non-uniform sampling than the baselines, highlighting the importance of avoiding quantization of the point cloud for high quality surface reconstruction.

\section{Conclusions}
Voxel-based data structures dominate surface reconstruction for large scale point clouds due to their superior time efficiency.
Despite its dominance, voxelization, which collapses multiple points within a cube into a single voxel feature, causes significant information loss, leading to degraded performance in the task of surface reconstruction.
In this work, we propose an efficient \textit{point-based} framework that allows us to use the original point cloud, hence not incurring in any information loss.
The key idea is point cloud serialization, inspired by recent efficient point transformers~\cite{wu2024point, wang2023octformer}, \at{coupled with a simple point-cloud architecture for feature aggregation.}
We show that our method outperforms prior SOTA in both accuracy and time efficiency, enabling point-based representation for surface reconstruction from large scale point clouds.  

\paragraph{Future works}
We would like to explore better strategies for combining the optimal KNN and fast approximate neighborhood function, as this strategy should ideally be adaptive to point cloud size.
We would also be interested in exploring generative modeling of large scale point clouds using these methods.

\section*{Acknowledgements}
This work was supported in part by the Natural Sciences and Engineering Research Council of Canada (NSERC) Discovery Grant, NSERC Collaborative Research and Development Grant, Google DeepMind, Digital Research Alliance of Canada, the Advanced Research Computing at the University of British Columbia, and the SFU Visual Computing Research Chair program.
Shaobo Xia was supported by National Natural Science Foundation of China under Grant 42201481.
We would also like to thank Jiahui Huang for the valuable discussion and feedback.

{
    \small
    \bibliographystyle{ieeenat_fullname}
    \bibliography{main}
}
\appendix
\clearpage
\setcounter{page}{1}
\maketitlesupplementary

This appendix provides additional ablation studies, experimental analyses and more qualitative results.
\section{Additional ablation studies}

\paragraph{Extending~\Cref{tab:agg_capacity} with more levels -- \Cref{tab:agg_capacity_supp}} 
A large non-linear aggregation module is essential to our method due to the false negatives in the fast approximate neighbors. However, more layers in the aggregation module degrades time efficiency. We show that $\aggregation$ with 2 non-linear layers suffices to achieve the best trade-off between accuracy and time efficiency.

\begin{table}[b]
\centering
\resizebox{.85\columnwidth}{!}{
\begin{tabular}{@{}lcccccc@{}}
\toprule
\makecell{\bf Num. of hidden\\\bf layers in $\aggregation$} & CD (10\textsuperscript{-2}) $\downarrow$ & F-score $\uparrow$ & Latency (s) $\downarrow$ \\ \midrule
 0 & 0.264 & 99.16 & 130 \\
 1 & 0.262 & 99.22 & 133 \\
 \rowcolor{1st}2 & 0.257 & 99.33 & 152 \\
 3 & 0.258 & 99.34 & 158 \\
 4 & 0.256 & 99.32 & 166 \\
 5 & 0.256 & 99.37 & 167 \\
\bottomrule
\end{tabular}
}
\caption{
{\bf Impact of capacity of $\aggregation$} -- the extension of \Cref{tab:agg_capacity} with more levels. The larger aggregation module achieves better performance with decreased time efficiency. We show that 2 layers achieves the best trade-off between accuracy and time efficiency. }

\label{tab:agg_capacity_supp}
\end{table}

\paragraph{Different ways to fuse per-scale features -- \Cref{tab:multi_level_agg}} 
We show the different ways to fuse the per-scale features and observe that they have similar accuracy. Attentive pooling achieves slightly better performance at the cost of degraded time efficiency. 
Note that we have an additional linear layer to predict the attention from the  concatenated features of all levels to perform the attentive pooling.
To realize a learnable gate where we multiply per-level weights with features before fusing levels, we train an additional learnable per-level weight followed by a Sigmoid function for the multiplication.
\begin{table}
\centering
\resizebox{.8\columnwidth}{!}{
\begin{tabular}{@{}lcccc@{}}
\toprule
\bf Fusion method & CD (10\textsuperscript{-2}) $\downarrow$ & F-score $\uparrow$ & Latency (s) $\downarrow$ \\ \midrule
\rowcolor{1st}Sum & 0.257 & 99.33 & 152 \\
Average & 0.257 & 99.33 & 151 \\
Concatenation & 0.256 & 99.37 & 151 \\
Learnable Gate & 0.257 & 99.33 & 152 \\
Attentive Pooling & 0.255 & 99.36 & 156 \\
\bottomrule
\end{tabular}
}
\caption{
{\bf Scales fusion} -- we investigate different ways to fuse per-scale features. Attentive pooling achieves marginal improvement at the cost of noticeable increased latency.}

\label{tab:multi_level_agg}
\end{table}

\section{More Experimental Analysis}
\begin{table}
\centering
\resizebox{.95\columnwidth}{!}{
\begin{tabular}{@{}lcccc@{}}
\toprule
\makecell{\bf Method} & CD ($10^{-2}$) $\downarrow$ & Peak Memory $(GB)$ $\downarrow$ & Latency/Iter. (s) $\downarrow$ \\ \midrule
\nksr & 0.246 & 41.3 & 1.44\\
\rowcolor{1st}Ours & 0.257 & 4.6 & 0.59 \\
Ours(w/KNN) & 0.243 & 8.7 & 0.64 \\
Ours (Minkowski) & 0.301 & 3.4 & 0.27 \\
\bottomrule
\end{tabular}
}
\caption{
{\bf Overhead during training} We report the overhead during training in terms of GPU peak memory and latency required for each training iteration. We show that our method achieves more efficient training than the current SOTA~\cite{huang2023neural}.}
\label{tab:training_speed_memory}
\end{table}

\paragraph{Overhead during training -- \Cref{tab:training_speed_memory}} 
We report our method's overhead during training in terms of GPU peak memory and latency required per each training iteration.   
Additionally, we profile training overhead (GPU peak memory and latency per iteration) on a single NVIDIA A6000 Ada with the PyTorch Lighting API. In all cases, we use a batch size of 1 for a fair comparison with the SOTA~\cite{huang2023neural} that has the batch size of 1 in one backward pass.  

\begin{table}
\centering
\resizebox{\columnwidth}{!}{
\begin{tabular}{@{}lccccc@{}}
\toprule
\makecell{\bf Methods} & \makecell{Feature \\ Backbone(s)} & Decoder(s) & \makecell{Dual Marching \\ Cube(s)} & Total (s) & CD (10\textsuperscript{-2}) $\downarrow$ \\ \midrule
\nksr & 83 & 313 & 78 & 480 & 0.246\\
\rowcolor{1st}Ours & 10 & 70 & 68 & 152 & 0.243\\
Ours (w/ KNN) & 10 & 72 & 68 & 151 & 0.257\\
Ours (Minkowski) & 6 & 30 & 56 & 97 & 0.301\\
\bottomrule
\end{tabular}
}
\caption{
{\bf Latency distribution.} 
We report the latency distribution during inference steps for the feature backbone $\backbone$, decoder and marching cubes. Our method outperforms the SOTA~\cite{huang2023neural} in all steps, particularly in the decoder step where~\cite{huang2023neural} needs to solve a large differentiable linear system.
}
\label{tab:detailed_time}
\end{table}

\paragraph{Latency distribution in the steps of inference  -- \Cref{tab:detailed_time}} 
We show that our method achieves better time efficiency than the SOTA~\cite{huang2023neural} in all different steps whilst having better accuracy, even without the time-consuming decoder as in SOTA.

\begin{figure}
  \centering
  \includegraphics[width=.9\columnwidth]{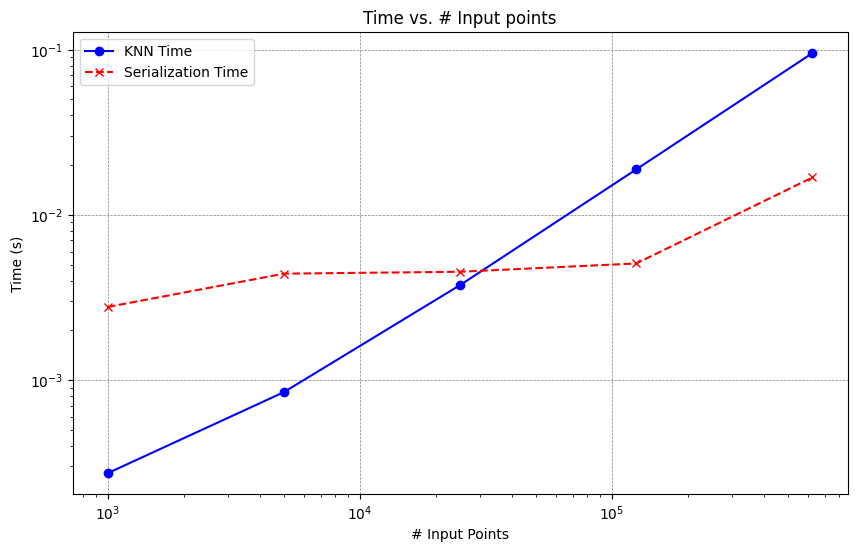}
\caption{{\bf Impact of point cloud size on time efficiency} We observe that K-nearest-neighbor (KNN) is more efficient than neighbors based on serialization encoding when the number of points is smaller than 25000. We suspect this is because KNN is highly optimized with a CUDA implementation, while the serialization encoding is purely based on Python.}
\label{fig:knn_serialization_npts}
\end{figure}

\paragraph{The impact of point cloud size on time efficiency of KNN vs serialization encoding -- \Cref{fig:knn_serialization_npts}}
We report how the point cloud size impacts the time efficiency of KNN and neighbors based the serialization encoding. Theoretically, serialization encoding should be more efficient. However, we observe that when the point cloud size is small such as \synthetic and \scannet, KNN is more efficient then serialization encoding. We suspect this is because KNN is highly engineered, with a CUDA implementation while the serialization encoding is purely implemented in python.  To estimate the time efficiency, we randomly generate 25000 query points and record the execution times of methods based on KNN and serialization encoding across varying numbers of input points.

\paragraph{More metrics: completeness and accuracy -- \Cref{tab:across_domain_supp} and \Cref{tab:indomain_supp}}
Following the state of the art method by~\citet{huang2023neural}, we further report additional metrics below.
We observe that the performance is consistent with other metrics we report in main paper.   

\begin{table}[]
\centering
\resizebox{.95\columnwidth}{!}{
\begin{tabular}{@{}lcccccc@{}}
\toprule
\makecell{\bf Num. of segments\\\bf in training and reconstruction} & CD (10\textsuperscript{-2}) $\downarrow$ & F-score $\uparrow$ & Latency (s) $\downarrow$ & Peak memory (GB) $\downarrow$\\ \midrule
\rowcolor{1st} 1 & 3.3 & 97.4 & 1.7 & 20.4 \\
 10 & 3.4 & 96.7 & 3.0 & 7.1 \\
 50 & 3.4 & 96.6 & 6.2 & 5.0 \\
\bottomrule
\end{tabular}
}
\caption{
{\bf Handling large scenes via partition -- } Simply with serialization codes, we partition a large scene into smaller segmentation to avoid GPU memory. We show that our method reduce the peak memory with a negligible decrease in reconstruction quality.}

\label{tab:segment_supp}
\end{table}

\paragraph{{Handling infinitely large scenes} --
\Cref{tab:segment_supp}}
\ws{We show that our method is capable of handling the infinitely large scenes. 
With serialization codes, we partition a large scene into segments and extract feature of segments individually, avoiding exploding the GPU memory. 
Note the partition stops message passing between segments, which harms the reconstruction accuracy. Even though, as shown in \Cref{tab:segment_supp}, our method achieves the good trade-off between reconstruction quality and the peak memory usage.}

\begin{table}[]
\centering
\resizebox{.60\columnwidth}{!}{
\begin{tabular}{@{}lcccccc@{}}
\toprule
\makecell{\bf weight. of $\loss{laplacian}$} & CD (10\textsuperscript{-2}) $\downarrow$ & F-score $\uparrow$ \\ \midrule
\rowcolor{1st} 0 & 3.3 & 97.4 \\
 1e-4 & 3.5 & 96.3 \\
 1e-3 & 3.6 & 95.8\\
\bottomrule
\end{tabular}
}
\caption{
{\bf Impact of the weights on Laplacian loss during training.} }
\label{tab:laplacian_supp}
\end{table}

\begin{figure}
  \centering
  \includegraphics[width=0.9\linewidth]{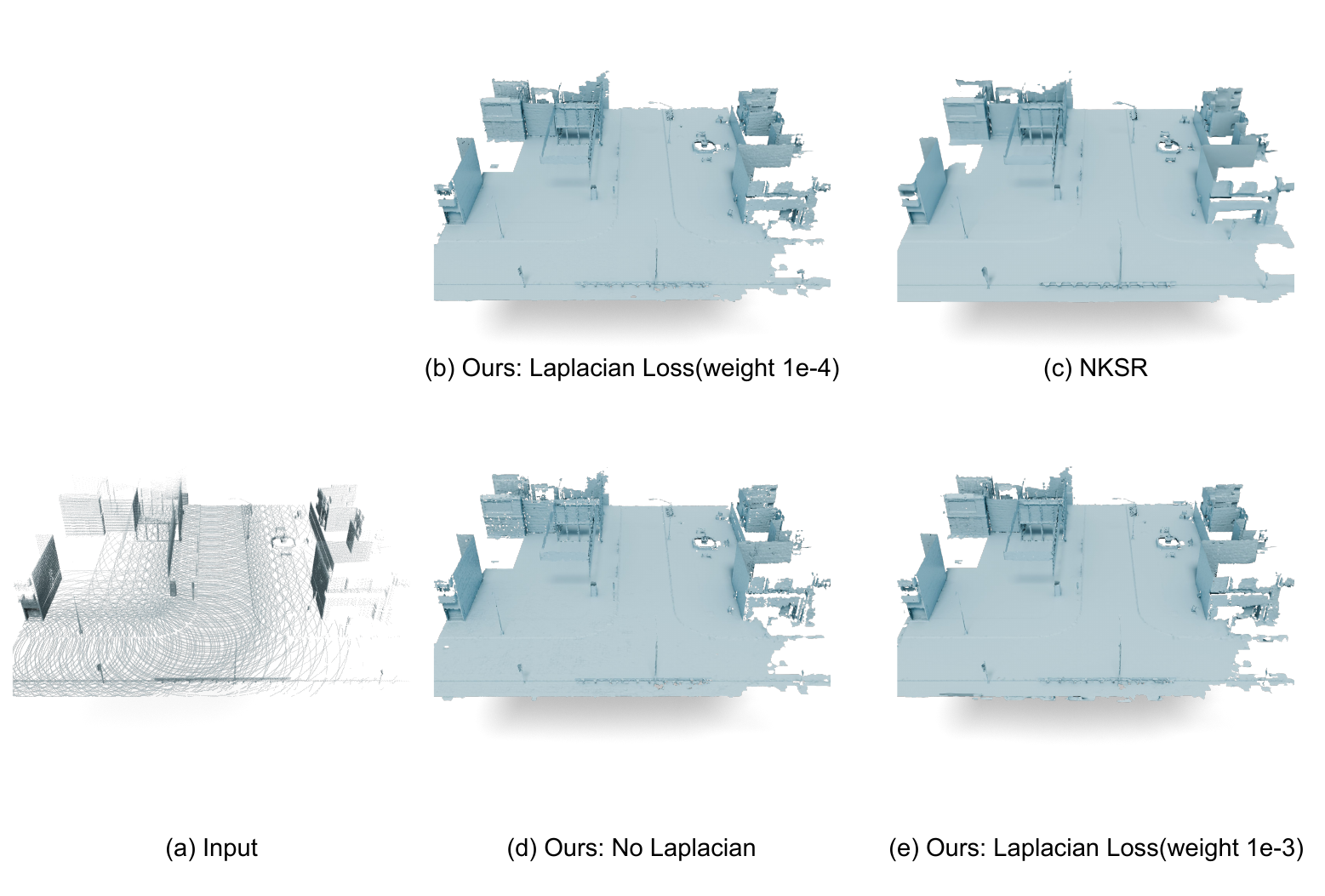}
\caption{{\bf \cl{ Smoother surface on \carla by Laplacian Loss }}}
\label{fig:laplacian_carla_supp}
\end{figure}

\paragraph{\ws{Smoother surfaces with Laplacian loss} --
\Cref{tab:laplacian_supp} and \Cref{fig:laplacian_carla_supp}}
\ws{We show that our method achieves smoother surfaces by regularizing the distance field $\distancefield$ with Laplacian loss from~\cite{benshabat2023digsdivergenceguided}. We define the loss as 
\begin{align}
    \loss{Laplacian} = \expect_{\x \sim \mathcal{Q}} \left[ \nabla^2 \distancefield(\x) \right].
\end{align}
As shown in \Cref{fig:laplacian_carla_supp}, the larger weight of $\loss{Laplacian}$ leads to the smoother surface. However, as a downside, the reconstruction accuracy is degraded as shown in \Cref{tab:laplacian_supp}. Nevertheless, with the weight of \(1 \times 10^{-4}\), our method achieves the better reconstruction accuracy than \nksr while having the similar surface smoothness. 
}

\begin{table}[]
\centering
\resizebox{.95\columnwidth}{!}{
\begin{tabular}{@{}lcccccc@{}}
\toprule
Methods & Primitive & CD (10\textsuperscript{-2}) $\downarrow$ & IoU $\uparrow$\\ \midrule
\rowcolor{1st} \nksr & Voxels & 2.34 & 95.6 \\
Ours~(Minkowski w/ KNN)~\cite{choy20194d} {(w/ KNN)}  & Voxels & 4.36 & 87.5 \\
Ours~(w/ KNN)& Points & 3.91 & 89.9 \\
Ours~(w/ KNN, w/ similar DMC grid number) & Points & 2.88 & 94.6 \\
\bottomrule
\end{tabular}
}
\caption{
{\bf Evaluation on ShapeNet~\cite{chang2015shapenet}}
}

\label{tab:shapenet_supp}
\end{table}

\paragraph{\ws{Performance on synthetic object-level dataset} --
\Cref{tab:shapenet_supp}}
\ws{We evaluate the reconstruction quality on ShapeNet\cite{chang2015shapenet}, a synthetic object-level dataset.
Note we use the data prepared by \nksr, and the smaller grid size (0.005) during serialization to avoid collisions.
As shown in \Cref{tab:shapenet_supp}, our method outperforms voxel-based methods, while performs worse than~\nksr.
We suspect that the ``voxel-growing'' strategy in \nksr is crucial to the synthetic object-level dataset, and we leave the integration of this strategy into our method for future work.}

\begin{table*}[]
\centering
\resizebox{\linewidth}{!}{
\setlength{\tabcolsep}{2pt}
\begin{tabular}{LcccccccccccccccccccccC}
\toprule
Methods & & \multicolumn{5}{c}{{\bf \synthetic}}  &  \multicolumn{5}{c}{{\bf \scannet}} & \multicolumn{5}{c}{\cl{{\bf \carla(Original)}}} & \multicolumn{5}{c}{{\bf \carla(Novel)}} \\
 \cmidrule(lr){3-7} \cmidrule(lr){8-12} \cmidrule(lr){13-17} \cmidrule(lr){18-22}
&Primitive& \makecell{CD \\ ($10^{-2}$) $\downarrow$} & \makecell{completeness\\($10^{-2}$)$\downarrow$} & \makecell{accuracy\\($10^{-2}$)$\downarrow$ }& F-Score  $\uparrow$ & \cl{ Latency (s) $\downarrow$ } & \makecell{CD \\ ($10^{-2}$) $\downarrow$ }& \makecell{completeness\\($10^{-2}$)$\downarrow$ }& \makecell{accuracy\\($10^{-2}$)$\downarrow$ }& F-Score  $\uparrow$ & \cl{Latency (s) $\downarrow$ } & CD (cm) $\downarrow$ & completeness(cm)$\downarrow$ & accuracy(cm)$\downarrow$ & F-Score  $\uparrow$ & \cl{Latency (s) $\downarrow$} & CD (cm) $\downarrow$ & completeness(cm)$\downarrow$ & accuracy(cm)$\downarrow$ & F-Score  $\uparrow$ & \cl{Latency (s) $\downarrow$} \\       
\midrule
SA-CONet~\cite{tang2021SACon} & Voxels & {0.496} & - & - & {93.60}  & - & - & - & - & - & - & - & - & - & - & - \\
ConvOcc~\cite{peng2020convoccnet} & Voxels & {0.420} & - & - & {96.40}  & - & - & - & -& - & - & - & - & - & - & - \\
NDF~\cite{chibane2020ndf} & Voxels & {0.408} & - & - & {95.20}  & - & 0.385 & - & - & 96.40  & -  & - & - & - & - & -\\
RangeUDF~\cite{wang2022rangeudf} & Voxels & {0.348}  & - & - & {97.80} & {-} & 0.286 & - & - & 98.80  & - & - & - & - & - & - \\
\cl{TSDF-Fusion~\cite{zeng20163dmatch}}  & - & - & - & - & -  & - & - & - & - & - & - & 8.1 & 8.0 & 8.2 & 80.2 & - & 7.6 & 6.6 & 8.6 & 80.7 & - \\
\cl{POCO~\cite{boulch2022poco}}  & - & - & - & - & - & - & - & - & - & - & - & 7.0 & 3.6 & 10.5 & 90.1 & - & 12.0 & 2.9 & 9.1 & 92.4 & - \\
\cl{SPSR~\cite{kazhdan2013screened}} & - & - & - & - & - & - & - & - & - & - & - & 13.3 & 16.4 & 10.3 & 86.5 & - & 11.3 & 12.8 & 9.9 & 88.3 & - \\
\nksr & Voxels & {0.345} & 0.304 & 0.387 & {97.26} & {0.40} & {0.246} & 0.221 & 0.27 & {99.51} & {1.54} & {3.9} & 2.2 & 5.6 & {93.9} & 2.0 & 2.8 & 2.1 & 3.6 & 96.0 & 1.8\\
\nksr (more data) & Voxels & - & - & - & - & - & - & - & - & - & - & {3.5} & {3.0} & {4.1} & {94.1} & 2.0 & 3.0 & 2.4 & 3.6 & 96.0 & 1.8\\
\makecell{Ours~(Minkowski)~\cite{choy20194d} \\ \scriptsize{(w/ KNN)} }& Voxels & - & \todo{} & \todo{} & \todo{} & \todo{} & 0.254 & 0.234 & 0.273 & 99.41 & 0.46 & 3.4 & 4.1 & 2.7 & 97.2 & 1.9 & 2.7 & 3.1 & 2.4 & 98.1 & 2.0 \\
Ours~(Minkowski)~\cite{choy20194d} & Voxels & - & \todo{} & \todo{} & \todo{} & \todo{} & 0.301 & 0.327 & 0.275 & 98.48 & 0.31 & 3.8 & 4.4 & 3.2 & 96.2 & 1.5 & 3.0 & 3.3 & 2.8 & 97.4 & 1.5\\
\rowcolor{1st} Ours \scriptsize{(w/ KNN)} & Points & {0.322} & {0.270} & {0.374} & {98.25} & {0.13} & {0.243} & {0.230} & {0.256} & {99.61} & {0.48} & {3.2} & {3.6} & {2.8} & {97.5} & 3.2 & 2.6 & 2.7 & 2.4 & 98.3 & 3.4\\
\rowcolor{1st}Ours & Points & 0.358 & 0.318 & 0.399 & 96.43 & 0.14 & 0.257 & 0.243 & 0.270 & 99.33 & 0.49 & 3.3 & 3.9 & 2.6 & 97.4 & 1.7 & 2.7 & 3.0 & 2.4 & 98.2 & 1.7 \\
\bottomrule
\end{tabular}
}
\caption{\cl{\textbf{Additional metrics from \nksr for in-domain evaluation}} 
}
\label{tab:indomain_supp}
\end{table*}

\begin{table*}[]
\centering
\resizebox{\linewidth}{!}{
\setlength{\tabcolsep}{4pt}
\begin{tabular}{LccccccccccccccccC}
\toprule
Methods & & \multicolumn{5}{c}{{\bf \synthetic $\rightarrow$ \scannet}}  &  \multicolumn{5}{c}{\cl{{\bf \scannet $\rightarrow$ \synthetic}}} & \multicolumn{5}{c}{\cl{{\bf \scannet $\rightarrow$ \scenenn}}} \\ 
 \cmidrule(lr){3-7} \cmidrule(lr){8-12} \cmidrule(lr){13-17}
&Primitive&\makecell{CD \\($10^{-2}$) $\downarrow$ }& \makecell{completeness\\($10^{-2}$)$\downarrow$ }& \makecell{accuracy\\ ($10^{-2}$)$\downarrow$ }& F-Score  $\uparrow$ & \cl{Latency (s) $\downarrow$ } & \makecell{CD \\ ($10^{-2}$) $\downarrow$ }& \makecell{completeness \\ ($10^{-2}$)$\downarrow$ }& \makecell{accuracy\\ ($10^{-2}$)$\downarrow$ }& F-Score  $\uparrow$ & \cl{Latency (s) $\downarrow$ } & \makecell{CD \\ ($10^{-2}$) $\downarrow$ }& \makecell{completeness\\ ($10^{-2}$)$\downarrow$ }& \makecell{accuracy\\ ($10^{-2}$)$\downarrow$ }& F-Score  $\uparrow$ & \cl{ Latency (s) $\downarrow$ }\\        
\midrule
SA-CONet~\cite{tang2021SACon} & Voxels & 0.845 & - & - & 77.80  & - & - & - & - & - & - & - & - & - & - & - \\
ConvOcc~\cite{peng2020convoccnet} & Voxels & 0.776 & - & - & 83.30  & - & - & - & - & -& - & - & - & - & - & - \\
NDF~\cite{chibane2020ndf} & Voxels & 0.452 & - & - & 96.00 & - & {0.568} & - & - & {88.10} & - & 0.425 & - & - & 94.80 & - \\
RangeUDF~\cite{wang2022rangeudf} & Voxels & {0.303} & - & - & {98.60} & {-} & 0.481 & - & - & 91.50 & - & 0.324 & - & - & 97.80 & - \\
\nksr & Voxels & {0.329} & {0.296} & {0.362} & {97.37} & {2.02} & {0.351} & {0.301} & {0.401} & {97.41} & {0.46} & {0.268}& {0.253} & {0.283} & {99.18} & {1.95} \\
\rowcolor{1st} Ours (w/ KNN) & Points & {0.284} & 0.266 & 0.302 & {98.65} & {0.54} & {0.327} & 0.263  & 0.391 & {98.37} & {0.13} & {0.277} & 0.277 & 0.277 & {99.00} & {0.50} \\
\bottomrule
\end{tabular}
}
\caption{\cl{\textbf{Additional metrics from \nksr for cross-domain evaluation} 
}}
\label{tab:across_domain_supp}
\end{table*}

\section{More qualitative results}
\begin{figure} 
  \centering
  \vspace{-1.2em}
  \includegraphics[width=\linewidth, trim={0 1.cm 0 0},clip]{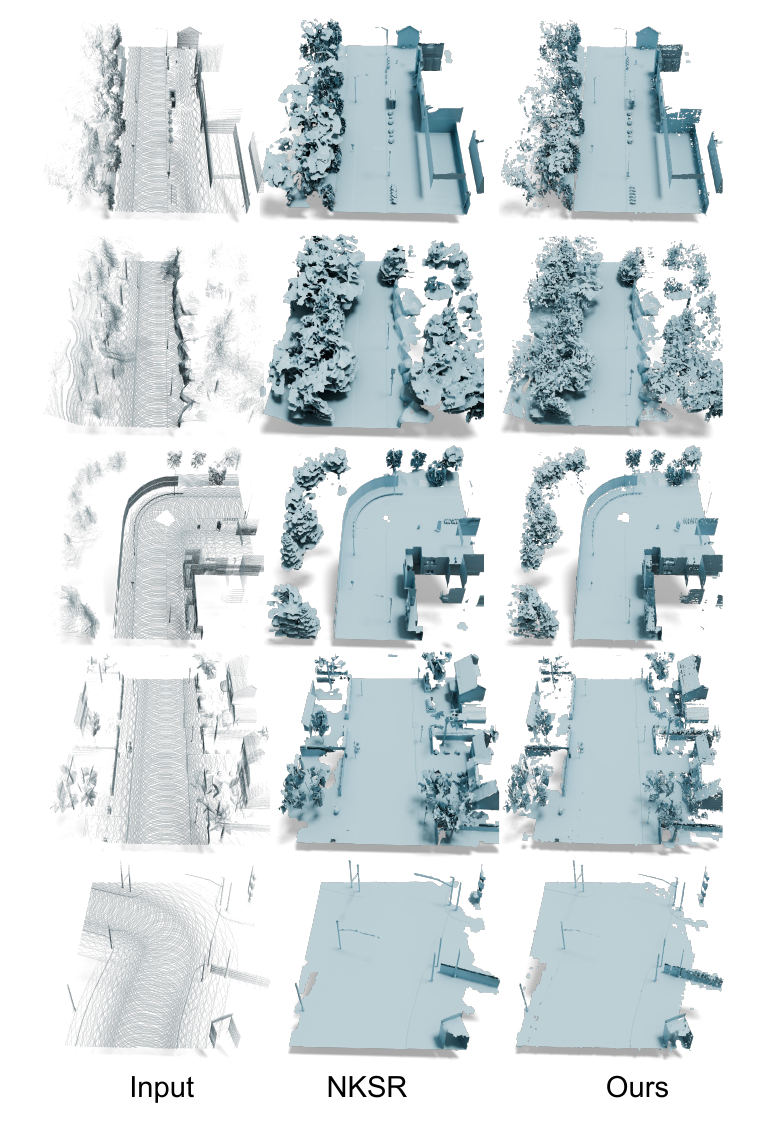}
    \begin{tblr}{colspec={X[c]X[c]X[c]}}
        \scriptsize{(a) Input point cloud} & \scriptsize{(b) \nksr} & \scriptsize{(c) Ours} \\
    \end{tblr} 
\caption{\ws{{\bf More qualitative results on \carla } -- Zoom in for better view.}}
\label{fig:carla_additional_1}
\end{figure}

\begin{figure}
  \centering
  \includegraphics[width=\linewidth, trim={0 1.cm 0 0},clip]{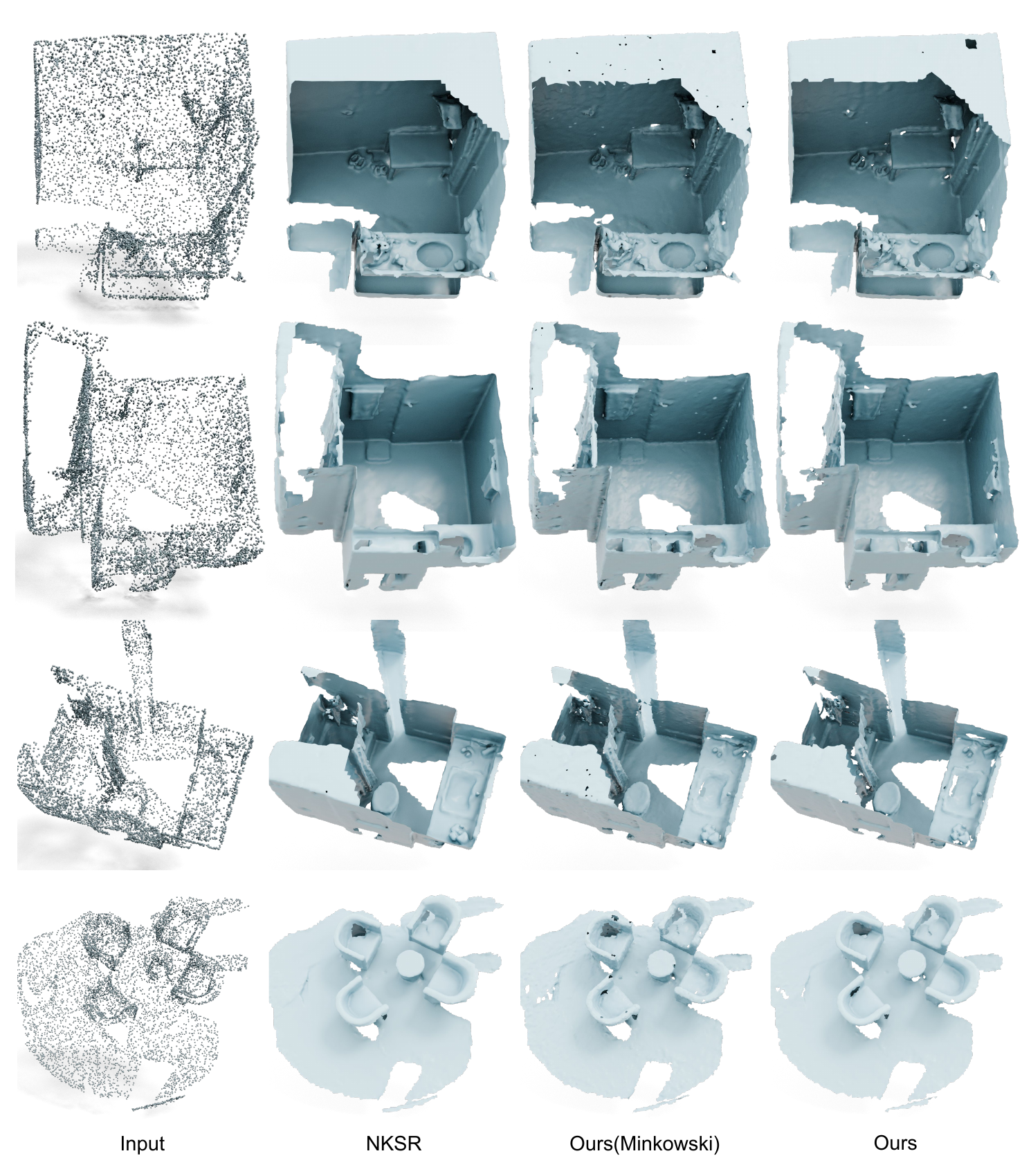}
    \begin{tblr}{colspec={X[c]X[c]X[c]X[c]}}
        \scriptsize{(a) Input point cloud} & \scriptsize{(b) \nksr} & \scriptsize{(c) Ours(Minkowski)} & \scriptsize{(d) Ours}\\
    \end{tblr} 
\caption{{\bf More qualitative results on \scannet} -- \ws{Zoom in for better view.}}
\label{fig:scannet_additional_1}
\end{figure}

\begin{figure}[t]
  \centering
  \includegraphics[width=\linewidth, trim={0 1.cm 0 0},clip]{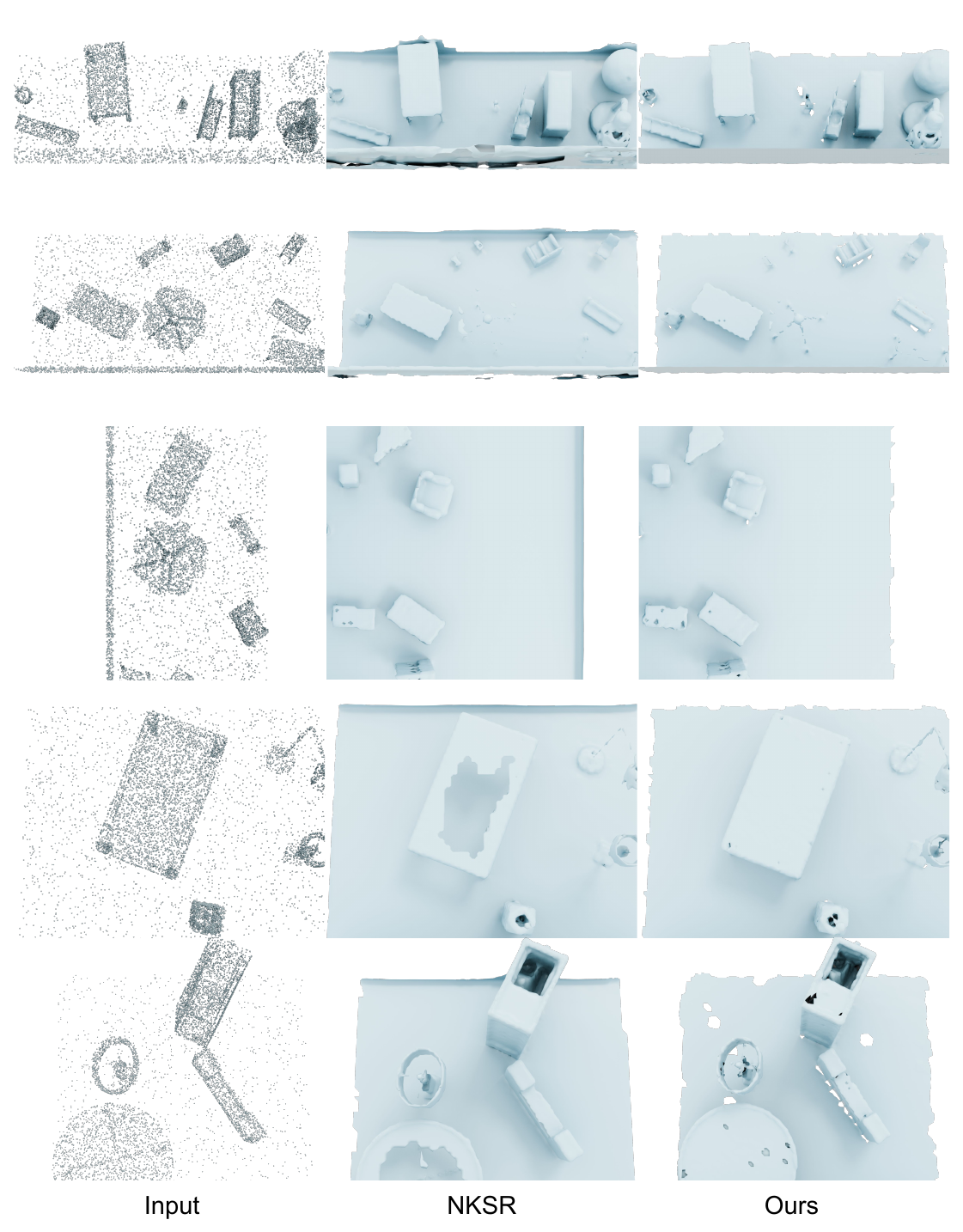}
    \begin{tblr}{colspec={X[c]X[c]X[c]}}
        \scriptsize{(a) Input point cloud} & \scriptsize{(b) \nksr} & \scriptsize{(c) Ours} \\
    \end{tblr} 
\caption{\ws{{\bf More qualitative results on \synthetic} -- Zoom in for better view.} }
\label{fig:synthetic_additional_1}
\end{figure}

We provide more qualitative results in \Cref{fig:carla_additional_1}, \Cref{fig:scannet_additional_1} and \Cref{fig:synthetic_additional_1}.

\end{document}